\documentclass[journal]{IEEEtran}
\usepackage[left=0.75in, right=0.75in, top=1in, bottom=0.9in]{geometry}
\usepackage{mathrsfs,amsthm}
\usepackage{cite}
\usepackage{amsmath}
\usepackage{algorithm}
\usepackage[noend]{algpseudocode}

\makeatletter
\def\BState{\State\hskip-\ALG@thistlm}
\def\ps@headings{%
    \def\@oddhead{\mbox{}\scriptsize\rightmark \hfil \thepage}%
    \def\@evenhead{\scriptsize\thepage \hfil \leftmark\mbox{}}%
    \def\@oddfoot{}%
    \def\@evenfoot{}}
\makeatother 
\usepackage{graphicx,amsmath,amssymb,stfloats,subfigure,multicol}
\usepackage{epsfig,epsf,psfrag,amssymb,amsfonts,latexsym,graphicx,mathrsfs,subfigure}
\usepackage{bbm}
\usepackage{chngpage}
\usepackage[dvips]{color}
\usepackage{textcomp}
\usepackage{hyperref}
\usepackage{url}
\usepackage[normalem]{ulem}
\usepackage{algpseudocode}
\newsavebox{\ieeealgbox}

\usepackage{array}

\def\old#1{}    

\def\nn{\nonumber}
\def\beq{\begin{equation}}
\def\eeq{\end{equation}}
\def\bea{\begin{eqnarray}}
\def\eea{\end{eqnarray}}
\def\ba{\begin{array}}
\def\ea{\end{array}}

\def\bitem{\begin{itemize}}
\def\eitem{\end{itemize}}
\def\ben{\begin{enumerate}}
\def\een{\end{enumerate}}

\definecolor{bgrd}{rgb}{1,1,1}
\definecolor{gray}{rgb}{0.5,0.5,0.5}
\definecolor{dkr}{rgb}{0.7,0.1,0.2}
\definecolor{dkb}{rgb}{0.1,0.1,0.8}

\makeatletter
\newdimen{\captionwidth}
\long\def\@makecaption#1#2{%
\captionwidth .9\hsize
\vskip 10pt%
\setbox\@tempboxa\hbox{#1: #2}%
  \ifdim \wd\@tempboxa >\captionwidth%
    \setbox\@tempboxa\hbox{#1:\hspace*{.5em}}%
    \hfil\parbox{\captionwidth}{\raggedright\hangindent \wd\@tempboxa%
    \hangafter=1\unhbox\@tempboxa#2}\hfill%
  \else\centerline{\box\@tempboxa}%
  \fi
}
\makeatother
\def\scalefig#1{\epsfxsize #1\textwidth}

\def\edoc{\end{document}}

\newcommand{\Zmsc}{\mathscr{Z}}

\def\Ec{{\cal E}}
\def\Fc{{\cal F}}

\def\Hc{{\cal H}}

\def\Nc{{\cal N}}

\def\Uc{{\cal U}}

\begin{document}

\title{Universal Data Anomaly Detection via Inverse Generative Adversary Network}
\author{Kursat Rasim Mestav,~\IEEEmembership{Student Member,~IEEE,} and Lang Tong,~\IEEEmembership{Fellow,~IEEE}
\thanks{\scriptsize
Kursat Rasim Mestav (\url{krm264@cornell.edu}) and Lang Tong (\url{lt35@cornell.edu}) are with the School of Electrical and Computer Engineering, Cornell University, Ithaca, NY 14853, USA. }
\thanks{\scriptsize Part of the work was presented at the 57th Allerton Conference on Communication, Control, and Computing \cite{Mestav:19Al}.}
\thanks{\scriptsize This work is supported in part by the National Science Foundation under
Award 1809830, 1932501, and, Power Systems and Engineering Research Center (PSERC) Research Project M-39}}
    
\maketitle

\begin{abstract}
    The problem of detecting data anomaly is considered.  Under the null hypothesis that models anomaly-free data, measurements are assumed to be from an unknown distribution with some authenticated historical samples.  Under the composite alternative hypothesis, measurements are from an unknown distribution positive distance away from the distribution under the null hypothesis. No training data are available for the distribution of anomaly data. A semi-supervised deep learning technique based on an inverse generative adversary network is proposed.
\end{abstract}


\section{Introduction} \label{sec:intro}
We consider the problem of detecting data anomaly under the following hypotheses. Under the null hypothesis $\Hc_0$  that models the anomaly-free data, measurements are from some unknown distribution $f_0$.  Under the alternative $\Hc_1$ that models anomaly, measurements are from an unknown distribution that is at least $\epsilon$ distance away from $f_0$. 

More precisely, given conditionally independent and identically distributed observations $Z_i, i=1,\cdots, N$, we consider the following hypothesis testing problem:
\beq \label{eq:H0H1}
\Hc_0: Z_i \sim f_0~~vs.~~\Hc_1: Z_i \sim f_1 \in \Fc,
\eeq
where $\Fc=\{f: ||f-f_0||> \epsilon\}$  and  $\|\cdot \|$ can be arbitrary distance measure such as the total variation or the KL divergence.

We refer the problem as universal anomaly detection for the reason that neither $f_0$ nor $f_1$ is known; nor do we assume that they belong to some known parametric families.  In the paradigm of data-driven solutions to anomaly detection, we assume instead that only a set of training samples $\Zmsc_0=\{z_{01},\cdots, z_{0T}\}$ under $\Hc_0$ is available.

The assumption that the alternative distribution is unknown reflects the fact that data anomaly can happen in many ways, including the possibility that an adversary may have tampered the data in a man-in-the-middle attack \cite{Conti:16CST}. Often in these cases, well-calibrated anomaly data are not available, or they are insufficient for learning.

The assumption that the distribution under the null hypothesis is unknown but with some training data is reasonable.  For instance, data may be measured under a quasi-stationary environment that some samples can be authenticated but not enough to estimate the distribution accurately. A data-driven approach to anomaly detection may prefer using training samples directly to construct a test rather than estimating $f_0$ first from the training data and using the estimated distribution to construct a test.

The  above hypothesis testing problem is general and has a wide range of applications in power system state estimation \cite{Mestav:19Al}, image processing \cite{Augusteijn:02IJRS}, and many others \cite{Pimentel:14SP}.

\subsection{Related Work}  There are limited results in the classical statistics and the statistical signal processing literature that treats the hypothesis testing problem above. Indeed, pathological examples exist that consistent detection may not even possible \cite{Pimentel:14SP}. The problem is nonparametric and lacks a specific structure that places the problem in a well-studied class.  The presence of training data under one hypothesis and the complete lack of training data in the other makes the problem a special machine learning problem.  Here we review some recent machine learning approaches.

In the machine learning literature, the above problem is considered as semi-supervised anomaly detection \cite{Chandola:2009ADS}. Some of the algorithms in this category can be classified into three groups: (i) the clustering-based and nearest neighborhood-based techniques, (ii) one-class support vector machine algorithm  and (iii) auto-encoder based neural-network approaches. 

Clustering-based methods such as \cite{Yuan:17IJCNN} rely on semi-supervised clustering, assuming that the anomaly-free data instances belong to a cluster whereas anomaly data do not. Nearest neighborhood-based techniques  \cite{Breunig:00SIGMOD} have a similar assumption that the distance of a data instance to its nearest neighbors is relatively small. Such topological assumptions may not be appropriate for anomalies that arise from data attacks where the attacker can manipulate data population. 

The technique of one-class SVM \cite{Scholkopf:99NIPS} learns a hyperplane to separate an anomaly-free region from the rest of the space. A kernel function can be used to generalize the technique for nonlinearly separable hypotheses.  For the universal data anomaly detection, choosing the right kernel function is highly nontrivial.

The auto-encoder based approaches \cite{Zhou:17ACM, Schlegl&Seebock:19} train an autoencoder on anomaly-free samples. The reconstruction errors of new samples are used as test statistics for anomaly detection. The work in  \cite{Schlegl&Seebock:19} is a GAN based autoencoder approach that uses a generator and the inverse of it together to construct an autoencoder. Such techniques do not perform well when the distributions of the anomaly and anomaly-free data overlap such that a well-trained auto-encoder cannot distinguish anomaly and anomaly-free distributions from which the data sample is drawn. 

The  proposed  technique  in  this  paper  builds  on  to  our work  focuses on a power system application \cite{Mestav:19Al}.  While both papers rely on the idea of a coincidence test, the proposed method in this paper (i) uses a different learning architecture for the inverse generative model, (ii)  uses a different learning algorithm for the inverse generative model, and (iii) proposes a new way of setting the threshold of the coincidence test. Also, we provide more substantial numerical tests, including the challenging problem of detecting unobservable data attacks.

\subsection{Summary of contributions}

\begin{figure}[ht]
\center
\begin{psfrags}
\psfrag{x}[c]{\small $Z_i$}
\psfrag{y}[c]{\small $Y_i$}
\psfrag{z}[l]{\small $X_i$}
\psfrag{G}[c]{\small $H$}
\psfrag{T}[c]{\small $K_1 \gtrless T_\alpha$}
\scalefig{0.5}\epsfbox{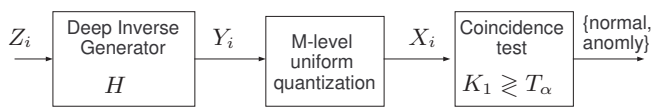}
\end{psfrags}
\caption{\small A schematic of universal data anomaly detection.}
\label{fig:scheme}
\end{figure}

We develop a novel anomaly detection approach consisting of an inverse generative adversarial network, a quantizer,  and a non-parametric coincidence test, illustrated in Fig.~\ref{fig:scheme}  The design of the three functional blocks is detailed in Sec.~\ref{UAD}.  

The key idea that allows us to distinguish the null hypothesis under $f_0$ from the alternative distributions in $\Fc$ is rooted in the classical birthday problem \cite{Mises:39}: given $M$ people in a room, what is the coincidence probability $P_c$  that there are at least two people having the same birthday?  

It turns out that this probability is the lowest when the underlying birthday distribution is uniform \cite{Mises:39}.  This suggests that a test on some measure of coincidence can serve as a way to distinguish the uniform distribution from all other distribution. Such a test was proposed earlier by David in \cite{David:50Biometrika} and more recently by Paniski \cite{Paninski:08TIT}. By thresholding, the number of unique people who do not share a birthday with others, the Paninski's test is shown to have both false alarm and miss-detection approaches to zero in the asymptotic regime.

The contribution of this work is to transform the problem of universal data anomaly detection to one of uniformity test for which consistent tests such as Paninski's coincidence test can be applied.  To this end, we employ an inverse generative adversary network (iGAN) as illustrated in Fig.~\ref{fig:scheme}.

Comparing with existing solutions, the proposed approach achieves diminishing detection error probabilities asymptotically assuming the (iGAN) is trained successfully. In the finite data sample regime, on the other hand, the proposed approach has low sample complexity in the sense that the number of testing samples is considerably smaller than the size of the quantization alphabet.  

We show through numerical examples that the proposed universal data anomaly detection algorithm is effective for some of the very challenging anomaly data scenarios, including the so-called unobservable attacks in power systems. 

\section{Universal Anomaly Detection}\label{UAD} 
\subsection{A  Schematic for Universal Anomaly Data Detection}
The idea of the proposed universal anomaly detection is captured in the schematic in Fig.~\ref{fig:scheme}.   Observation samples $\{Z_i\}$ are passed through an {\em inverse generator} $H$ that maps $Z_i \sim f_0$ to uniformly distributed samples $Y_i \sim \Uc(0,1)$ in interval $(0,1)$.   The existence of such a mapping is guaranteed by the fact that the cumulative distribution function $F_Z(\cdot)$ of $Z_i$ is one but not the only one such mapping.   Because $f_0$ is unknown,  the mapping is to be learned from the available historical data as shown in Sec.~\ref{IGAN}.

Upon successful training of the inverse generator $H$, under $H_0$, analog samples $Y_i$ are approximately i.i.d. uniformly distributed. They are then quantized uniformly with $M$ levels, which results in $M$-alphabet discrete uniformly distributed samples $X_i \sim \Uc(M)$.

A coincidence test using 1-coincidence statistic $K_1(x)$ produces the test outcome. The threshold is set depending on the level of acceptable false-alarm (the size) of the detector, the quantization level $M$, the number of test samples $N$, and the detection resolution $\epsilon$.  See Sec.~\ref{threshold}.

\subsection{Inverse Generative Adversary Network}\label{IGAN}  
\begin{figure}[ht]
\center
\begin{psfrags}
\psfrag{x}[r]{$Z \in \Zmsc_0 $}
\psfrag{u}[r]{$U \sim \Uc(0,1)$}
\scalefig{0.4}\epsfbox{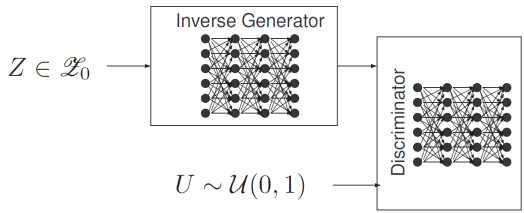}
\end{psfrags}
\caption{\small An inverse generative adversary network (WIGAN) learning of an inverse generator. }
\label{fig:IGAN}
\end{figure}

We propose the Wasserstein inverse generative adversary network (WIGAN) to produce an inverse generative model. Shown in Fig.~\ref{fig:IGAN}, WIGAN is a  modification of WGAN \cite{Arjovsky:17Arxiv}. We use the 1-Wasserstein distance to measure the similarity between probability distributions. In \cite{Arjovsky:17Arxiv} it is demonstrated that the Wasserstein distance is a more meaningful loss metric that leads to an improved stability of the optimization process.

WIGAN consists of two simultaneously trained neural networks: (i) an inverse generator and (ii) a discriminator. The training data passes through the inverse generator and the output is tested against synthetic uniformly distributed data by a discriminator. Ideally, the inverse generator converges to a function that transforms the distribution of the data to the uniform distribution.

An implementation of WIGAN is shown in  Algorithm 1. In our approach, the weights in both networks $\theta_{I},\theta_{D}$ are initialized randomly and updated with the learning rate of $\alpha$. To enforce the Lipschitz constraint of the 1-Wasserstein distance we used weight clipping with parameter $c$ on discriminator's updates as it is used in \cite{Arjovsky:17Arxiv}. The discriminator $f_{\theta_{D}}$ is updated more frequently  than the inverse generator $g_{\theta_{I}}$. We used RMSProp algorithm \cite{Tieleman:12Coursera} for the weight updates.

\begin{algorithm}
\caption{WIGAN. The experiments in the paper used the default values $\alpha = 0.001$, $c = 0.01$, $m = 100$, $n = 10$.}\label{WIGAN}
\begin{algorithmic}[1]
\Require: $\alpha$, the learning rate. $c$, the clipping parameter. $m$, the batch size. $n$, the number of iterations of the discriminator per generator iteration.
\For{Number of training iterations}
\For{$t = 0,1,...,n$}
\State Sample $\{U_{i}\}_{i=1}^m \sim \Uc(0,1)$ from uniform distribution.
\State Sample $\{Z_{i}\}_{i=1}^m \sim f_{0}$ from real data.
\State Update the discriminator parameters $\theta_{D}$ by descending its stochastic gradient:

$\nabla_{\theta_{D}} \big[ \frac{1}{m}\sum\limits_{i=1}^{m}{f_{\theta_{D}}(U_{i})}-\frac{1}{m}\sum\limits_{i=1}^{m}{f_{\theta_{D}}(g_{\theta_{I}}(Z_{i})}) \big]$
\State $\theta_{D} \gets clip(\theta_{D},-c,c)$
\EndFor
\State Sample $\{Z_{i}\}_{i=1}^m \sim f_{0}$ from real data .
\State Update the inverse generator parameters $\theta_{I}$ by descending its stochastic gradient:

$\nabla_{\theta_{I}} \big[ \frac{1}{m}\sum\limits_{i=1}^{m}{f_{\theta_{D}}(g_{\theta_{I}}(Z_{i})} \big]$
\EndFor
\end{algorithmic}
\end{algorithm}
\vspace{-0.4cm}

\subsection{Quantization and Coincidence Test}

Once the inverse generator is learned, we have the transformed data samples $Y_{i}$ that are uniformly distributed under $\Hc_0$ and nonuniform under $\Hc_1$.  Testing the uniformity of continuously distributed random samples without any assumptions on the density function is nontrivial \cite{Adamaszek:10book}.  Here we apply the $M$-level uniform quantization to $Y_i$, which gives us $M$-ary discrete random samples $X_i$ that are uniformly distributed under $\Hc_0$.  The distribution of $X_i$ under $\Hc_1$ depends on the hyper-parameter $M$, however. Whereas finding the optimal choice of $M$ is beyond the scope of this paper, we assume, for now, that almost everywhere in $\Fc$, the inverse transformed and quantized samples $X_i$ is $\epsilon$ distance away from being uniform. 

At the heart of the proposed approach is the coincidence test for uniformity proposed by Paninski \cite{Paninski:08TIT} for the following binary hypotheses using conditionally IID samples $\{X_i,i=1,\cdots, N\}$ from $M$-alphabet discrete distributions
\bea
&\Hc_0':& X_i \sim P_0=(\frac{1}{M},\cdots,\frac{1}{M}),\nn\\
&\Hc_1':& X_i \sim P_1 \in \{p=(p_1,\cdots, p_M)|~ ||p-P_0||>\epsilon\}.\nn
\eea

The intuition of uniformity test is that, when $X_i$ are from the uniform distribution, the probability of coincidence is the lowest, and $K_1(x)$, the number of ``unique'' valued samples, is the highest.  Thus, Paninski's test for uniformity is given by

\[
K_1(x)
\begin{array}{c}
{\small \Hc_0'}\\
\gtrless\\
{\small \Hc_1'}\\
\end{array}
 T_\alpha
\]
where the threshold $T_\alpha$ is a function of false positive level $\alpha$ as well as the alphabet size (quantization level) $M$, the sample size $N$, and distance between two hypotheses $\epsilon$.

Paninski showed that the coincidence test is consistent so long as $N$ grows faster than $\sqrt{M}$ as  $N = o(\frac{1}{\epsilon^4} \sqrt{M})$. Remarkably, the sample complexity can be significantly less than the size of the alphabet. A large-deviation bound is later established in \cite{Huang:13TIT}.

\subsection{Test threshold}\label{threshold}
When the sample size $N$ is finite, choosing the right threshold affects the true and false positive probabilities of the detection.  For the $K_1(x)$ test, setting the test threshold amounts to evaluating the probability of the event that $K_1(x) \ge t$.  

Let $P_0(\Ec)$ be the probability of event $\Ec$ under hypothesis $\Hc_0$.  The  threshold $T_\alpha$ of the $K_1$ coincidence test with the constraint on the false-positive probability to no greater than $\alpha$ is given by

\bea
T_\alpha = \min\{t: P_0(K_1 \le t) \le \alpha \}.
\eea
The computation of $T_\alpha$ amounts to evaluating $P_0(K=1)$, which was given by Von Mises in \cite{Mises:39}:
\[
P_0(K_1=k)=\sum_{j=k}^{M}(-1)^{j+k} {j\choose k} {m\choose j}\frac{ N!}{(N-j)!}\frac{(M-j)^{N-j}}{M^{N}}.
\]

\section{Simulation}
We present two sets of simulation results.  The first simulation is based on a synthetic data set generated from the two hypotheses. We used a composite hypothesis for the alternative hypothesis to capture the variability of the alternative hypotheses. 

The second simulation is about the detection of what is considered unobservable attack in power system state estimation. Such attacks are crafted in such a way that all existing techniques fail.

\subsection{Synthetic data}
We tested the proposed method on Gaussian and Gaussian Mixture models. We evaluated it on 2 scenarios,

Case 1: $\Hc_0: Z_i \sim \Nc(0,1)~{\rm\it v.s.}~\Hc_1: Z_i \sim \Nc(\mu,1)$ where $-1 < \mu < 1$.

Case 2: $\Hc_0: Z_i \sim \Nc(0,1)~{\rm\it v.s.}~\Hc_1: Z_i \sim \Nc(0,\sigma)$ where $0.5 < \sigma < 0.8$.

We used 10000 anomaly-free training samples to train the iWGAN. To test our algorithm, we generated 20000 batches of $N = 50$ samples from the distribution in $\Hc_0$ and in $\Hc_1$. For each batch, we varied the $\mu$ and $\sigma$. After using the iWGAN, we simply used a fixed value of 200 for the quantization parameter $M$ for all experiments. However, there is a space for improvement by choosing $M$ more judiciously. 

For each case, we compared the proposed approach with two major deep learning benchmarks: the autoencoder approach based on the reconstruction error of f-AnoGAN \cite{Schlegl&Seebock:19} and the One-Class SVM \cite{Scholkopf:99NIPS}. We implemented all methods on Python using the scikit-learn library \cite{Pedregosa:11MLR} and TensorFlow \cite{Abadi:15tensor}.

f-AnoGAN is based on the reconstruction error of the autoencoder. It assumes to have lower reconstruction error for anomaly-free data as it is trained using them. However, when the support of the  distribution of anomaly and anomaly-free data overlap such as Case 1 this assumption is not true. 

\begin{figure}[ht]
    \centering
    \includegraphics[width=1\columnwidth]{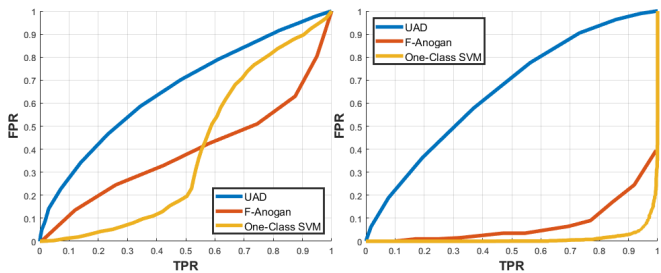}
    \hphantom{lol}
    \vspace{-0.5cm}
    \caption{\small ROC curves of the methods. Left: Case 1. Right: Case 2.}
    \label{fig:Case1and2}
\end{figure}

We trained the One-Class SVM using Radial Basis Function (RBF) as it had the best performance among the popular kernel choices. One-Class SVM tests new samples according to their closeness to the center of training samples. Similarly, in Case 1 this method performed poorly. To see a more extreme scenario, in Case 2 we tested a case where the anomaly samples are much denser around the mean. Both alternative methods performed significantly bad in that case, where UAD was still reasonable. The ROC curve of the detectors are presented in Fig.~\ref{fig:Case1and2}. 
\vspace{-0.2cm}

\vspace{-0.5cm}

\subsection{Detection of unobservable attack}

We present an application of the universal data anomaly detection to perhaps one of the most challenging  adversarial data attack detection problems. 

Consider a system with state vector $x$ and measurement $z$ satisfying 
\[
z=h(x)+e
\]
under the null hypothesis $\Hc_0$. We assume that we have historical measurements under $H_0$.

We assume the strongest attacker who has full access to system measurement function and system state. Suppose that the adversary can inject the so-called unobservable attack $a(x)=h(x+c)-h(x)$ into the measurement so that the system control center observes 
\[
z=h(x)+e+a(x)=h(x+c)+e,
\]
which means that the control center is deceived to believe that the actual state is $x'=x+c$.

Despite that the above attack appears to be unobservable by any algebraic means, the attack vector $a(x)$ does change the underlying distribution of $z$, which is where the proposed universal detection scheme can be effective in detecting such an unobservable attack. 

We simulated the unobservable attack on the IEEE 14 bus transmission test system \cite{Christie:14bus} using the load values in EPFL smart grid data \cite{Pignati&etal:15PESISG}. We designed an unobservable data attack by corrupting two of the measurements. We implemented ---the $J(x)$-test based on the classical $\xi^2$ test---and the deep learning-based anomaly detection methods. We used 10000 data samples to train the algorithms and 10000 batches of $N = 50$ samples from the attack-free samples and the samples with unobservable data attack to test.

\begin{figure}[ht]
    \centering
    \includegraphics[width=0.5\columnwidth]{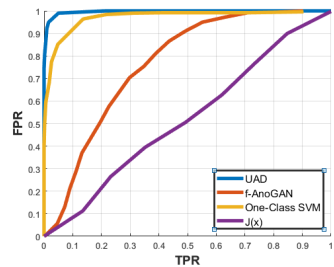}
    \hphantom{lol}
    \vspace{-0.2cm}
    \caption{\small ROC curve for the unobservable attack case.}
    \label{fig:Power}
\end{figure}

The $J(x)$ could not detect the data attacks as they were designed to be unobservable. However, the UAD detected it by monitoring the changes on the distribution of samples. One-Class SVM and f-AnoGAN also detected the bad data as they are trained using the anomaly-free data, but they did not perform as well as UAD possibly because the real data did not satisfy their assumptions. The ROC curve is presented in Fig.~\ref{fig:Power}.

\vspace{-0.2cm}
\section{Conclusion}\label{sec:conclusion}
This paper presents a novel method for the problem of detecting data anomaly under semi-supervised settings. The proposed method is an extension of the coincidence uniformity test using deep generative adversary networks. The proposed algorithm uses a direct approach to test the samples without explicitly learning the distribution that makes it is possible to have a decision using fewer samples. Numerical  tests  show considerable  gain  over  the  state of the art anomaly detection methods. 
    
{
    \bibliographystyle{IEEEtran}
    \bibliography{BIB}

\begin{thebibliography}{10}
\providecommand{\url}[1]{#1}
\csname url@samestyle\endcsname
\providecommand{\newblock}{\relax}
\providecommand{\bibinfo}[2]{#2}
\providecommand{\BIBentrySTDinterwordspacing}{\spaceskip=0pt\relax}
\providecommand{\BIBentryALTinterwordstretchfactor}{4}
\providecommand{\BIBentryALTinterwordspacing}{\spaceskip=\fontdimen2\font plus
\BIBentryALTinterwordstretchfactor\fontdimen3\font minus
  \fontdimen4\font\relax}
\providecommand{\BIBforeignlanguage}[2]{{%
\expandafter\ifx\csname l@#1\endcsname\relax
\typeout{** WARNING: IEEEtran.bst: No hyphenation pattern has been}%
\typeout{** loaded for the language `#1'. Using the pattern for}%
\typeout{** the default language instead.}%
\else
\language=\csname l@#1\endcsname
\fi
#2}}
\providecommand{\BIBdecl}{\relax}
\BIBdecl

\bibitem{Mestav:19Al}
K.~R. {Mestav} and L.~{Tong}, ``Learning the unobservable: High-resolution
  state estimation via deep learning,'' in \emph{2019 57th Annual Allerton
  Conference on Communication, Control, and Computing (Allerton)}, Sep. 2019,
  pp. 171--176.

\bibitem{Conti:16CST}
M.~Conti, N.~Dragoni, and V.~Lesyk, ``A survey of man in the middle attacks,''
  \emph{IEEE Communications Surveys Tutorials}, vol.~18, no.~3, pp. 2027--2051,
  thirdquarter 2016.

\bibitem{Augusteijn:02IJRS}
\BIBentryALTinterwordspacing
M.~F. Augusteijn and B.~A. Folkert, ``Neural network classification and novelty
  detection,'' \emph{International Journal of Remote Sensing}, vol.~23, no.~14,
  pp. 2891--2902, 2002. [Online]. Available:
  \url{https://doi.org/10.1080/01431160110055804}
\BIBentrySTDinterwordspacing

\bibitem{Pimentel:14SP}
\BIBentryALTinterwordspacing
M.~A. Pimentel, D.~A. Clifton, L.~Clifton, and L.~Tarassenko, ``A review of
  novelty detection,'' \emph{Signal Processing}, vol.~99, pp. 215 -- 249, 2014.
  [Online]. Available:
  \url{http://www.sciencedirect.com/science/article/pii/S016516841300515X}
\BIBentrySTDinterwordspacing

\bibitem{Chandola:2009ADS}
\BIBentryALTinterwordspacing
V.~Chandola, A.~Banerjee, and V.~Kumar, ``Anomaly detection: A survey,''
  \emph{ACM Comput. Surv.}, vol.~41, no.~3, pp. 15:1--15:58, Jul. 2009.
  [Online]. Available: \url{http://doi.acm.org/10.1145/1541880.1541882}
\BIBentrySTDinterwordspacing

\bibitem{Yuan:17IJCNN}
G.~{Yuan}, B.~{Li}, Y.~{Yao}, and S.~{Zhang}, ``A deep learning enabled
  subspace spectral ensemble clustering approach for web anomaly detection,''
  in \emph{2017 International Joint Conference on Neural Networks (IJCNN)}, May
  2017, pp. 3896--3903.

\bibitem{Breunig:00SIGMOD}
\BIBentryALTinterwordspacing
M.~M. Breunig, H.-P. Kriegel, R.~T. Ng, and J.~Sander, ``Lof: Identifying
  density-based local outliers,'' \emph{SIGMOD Rec.}, vol.~29, no.~2, pp.
  93--104, May 2000. [Online]. Available:
  \url{http://doi.acm.org/10.1145/335191.335388}
\BIBentrySTDinterwordspacing

\bibitem{Scholkopf:99NIPS}
\BIBentryALTinterwordspacing
B.~Sch\"{o}lkopf, R.~Williamson, A.~Smola, J.~Shawe-Taylor, and J.~Platt,
  ``Support vector method for novelty detection,'' \emph{Proceedings of the
  12th International Conference on Neural Information Processing Systems}, pp.
  582--588, 1999. [Online]. Available:
  \url{http://dl.acm.org/citation.cfm?id=3009657.3009740}
\BIBentrySTDinterwordspacing

\bibitem{Zhou:17ACM}
\BIBentryALTinterwordspacing
C.~Zhou and R.~C. Paffenroth, ``Anomaly detection with robust deep
  autoencoders,'' \emph{Proceedings of the 23rd ACM SIGKDD International
  Conference on Knowledge Discovery and Data Mining}, pp. 665--674, 2017.
  [Online]. Available: \url{http://doi.acm.org/10.1145/3097983.3098052}
\BIBentrySTDinterwordspacing

\bibitem{Schlegl&Seebock:19}
\BIBentryALTinterwordspacing
T.~Schlegl, P.~Seeböck, S.~M. Waldstein, G.~Langs, and U.~Schmidt-Erfurth,
  ``f-anogan: Fast unsupervised anomaly detection with generative adversarial
  networks,'' \emph{Medical Image Analysis}, vol.~54, pp. 30 -- 44, 2019.
  [Online]. Available:
  \url{http://www.sciencedirect.com/science/article/pii/S1361841518302640}
\BIBentrySTDinterwordspacing

\bibitem{Mises:39}
R.~Von~Mises, ``{\"U}ber aufteilungs-und besetzungswahrscheinlichkeiten,''
  \emph{Revue de la Faculté des Sciences de l’Université d’Istanbul},
  vol.~4, p. 145–163, 1939.

\bibitem{David:50Biometrika}
\BIBentryALTinterwordspacing
F.~N. David, ``Two combinatorial test of whether a sample has come from a given
  population,'' \emph{Biometrika}, vol.~37, no. 1/2, pp. 97--110, 1950.
  [Online]. Available: \url{http://www.jstor.org/stable/2332152}
\BIBentrySTDinterwordspacing

\bibitem{Paninski:08TIT}
L.~{Paninski}, ``A coincidence-based test for uniformity given very sparsely
  sampled discrete data,'' \emph{IEEE Transactions on Information Theory},
  vol.~54, no.~10, pp. 4750--4755, Oct 2008.

\bibitem{Arjovsky:17Arxiv}
M.~Arjovsky, S.~Chintala, and L.~Bottou, ``Wasserstein gan,'' 2017.

\bibitem{Tieleman:12Coursera}
T.~Tieleman and G.~Hinton, ``Lecture 6.5-rmsprop: Divide the gradient by a
  running average of its recent magnitude,'' \emph{COURSERA: Neural networks
  for machine learning}, vol.~4, no.~2, pp. 26--31, 2012.

\bibitem{Adamaszek:10book}
\BIBentryALTinterwordspacing
M.~Adamaszek, A.~Czumaj, and C.~Sohler, \emph{Testing Monotone Continuous
  Distributions on High-Dimensional Real Cubes}.\hskip 1em plus 0.5em minus
  0.4em\relax Berlin, Heidelberg: Springer Berlin Heidelberg, 2010, pp.
  228--233. [Online]. Available:
  \url{https://doi.org/10.1007/978-3-642-16367-8_13}
\BIBentrySTDinterwordspacing

\bibitem{Huang:13TIT}
D.~{Huang} and S.~{Meyn}, ``Generalized error exponents for small sample
  universal hypothesis testing,'' \emph{IEEE Transactions on Information
  Theory}, vol.~59, no.~12, pp. 8157--8181, Dec 2013.

\bibitem{Pedregosa:11MLR}
F.~Pedregosa, G.~Varoquaux, A.~Gramfort, V.~Michel, B.~Thirion, O.~Grisel,
  M.~Blondel, P.~Prettenhofer, R.~Weiss, V.~Dubourg, J.~Vanderplas, A.~Passos,
  D.~Cournapeau, M.~Brucher, M.~Perrot, and E.~Duchesnay, ``Scikit-learn:
  Machine learning in {P}ython,'' \emph{Journal of Machine Learning Research},
  vol.~12, pp. 2825--2830, 2011.

\bibitem{Abadi:15tensor}
\BIBentryALTinterwordspacing
M.~Abadi, A.~Agarwal, P.~Barham, E.~Brevdo, Z.~Chen, C.~Citro, G.~S. Corrado,
  A.~Davis, J.~Dean, M.~Devin, S.~Ghemawat, I.~Goodfellow, A.~Harp, G.~Irving,
  M.~Isard, Y.~Jia, R.~Jozefowicz, L.~Kaiser, M.~Kudlur, J.~Levenberg,
  D.~Man\'{e}, R.~Monga, S.~Moore, D.~Murray, C.~Olah, M.~Schuster, J.~Shlens,
  B.~Steiner, I.~Sutskever, K.~Talwar, P.~Tucker, V.~Vanhoucke, V.~Vasudevan,
  F.~Vi\'{e}gas, O.~Vinyals, P.~Warden, M.~Wattenberg, M.~Wicke, Y.~Yu, and
  X.~Zheng, ``{TensorFlow}: Large-scale machine learning on heterogeneous
  systems,'' 2015, software available from tensorflow.org. [Online]. Available:
  \url{http://tensorflow.org/}
\BIBentrySTDinterwordspacing

\bibitem{Christie:14bus}
R.~Christie, ``Power systems test case archive,'' \emph{Univ. Washington,
  Seattle, WA, USA.}, 1993,
  \url{https://labs.ece.uw.edu/pstca/pf14/pg_tca14bus.htm}.

\bibitem{Pignati&etal:15PESISG}
M.~{Pignati}, M.~{Popovic}, S.~{Barreto}, R.~{Cherkaoui}, G.~{Dario Flores},
  J.~{Le Boudec}, M.~{Mohiuddin}, M.~{Paolone}, P.~{Romano}, S.~{Sarri},
  T.~{Tesfay}, D.~{Tomozei}, and L.~{Zanni}, ``Real-time state estimation of
  the epfl-campus medium-voltage grid by using pmus,'' \emph{2015 IEEE Power
  Energy Society Innovative Smart Grid Technologies Conference (ISGT)}, pp.
  1--5, Feb 2015.

\end{thebibliography}
}

\end{document}